\documentclass{article}

\usepackage[final,nonatbib]{nips_2017}

\usepackage[utf8]{inputenc} % allow utf-8 input
\usepackage[T1]{fontenc}    % use 8-bit T1 fonts
\usepackage{hyperref}       % hyperlinks
\usepackage{url}            % simple URL typesetting
\usepackage{booktabs}       % professional-quality tables
\usepackage{amsfonts}       % blackboard math symbols
\usepackage{nicefrac}       % compact symbols for 1/2, etc.
\usepackage{microtype}      % microtypography
\usepackage{graphicx}
\usepackage{amsmath}
\usepackage{wrapfig}
\usepackage{subcaption}
\usepackage[numbers]{natbib}
\usepackage[toc,page]{appendix}
\usepackage{xr}
\usepackage{rotating}
\title{An Error Detection and Correction Framework for Connectomics}
\author{
	Jonathan Zung\\
	Princeton University\\
	{\tt jzung@princeton.edu}
	\And
	Ignacio Tartavull\thanks{Current address: Uber Advanced Technology Group}\\
	Princeton University\\
	{\tt tartavull@princeton.edu}
	\And
	Kisuk Lee\thanks{Co-affiliation: Princeton University}\\
	MIT\\
	{\tt kisuklee@mit.edu}
	\And
    H. Sebastian Seung\\
	Princeton University\\
	{\tt sseung@princeton.edu}
}
\begin{document}

\maketitle

\begin{abstract}
We define and study error detection and correction tasks that are
useful for 3D reconstruction of neurons from electron microscopic
imagery, and for image segmentation more generally. Both tasks take as
input the raw image and a binary mask representing a candidate
object. For the error detection task, the desired output is a map of
split and merge errors in the object. For the error correction task,
the desired output is the true object. We call this object mask
pruning, because the candidate object mask is assumed to be a superset
of the true object. We train multiscale 3D convolutional networks to
perform both tasks. We find that the error-detecting net can achieve
high accuracy. The accuracy of the error-correcting net is enhanced if
its input object mask is ``advice'' (union of erroneous objects) from
the error-detecting net.
\end{abstract}

\begin{figure}[t!]
	\begin{center}
	\begin{subfigure}[t]{0.45\textwidth}
		\includegraphics[width=1.0\linewidth]{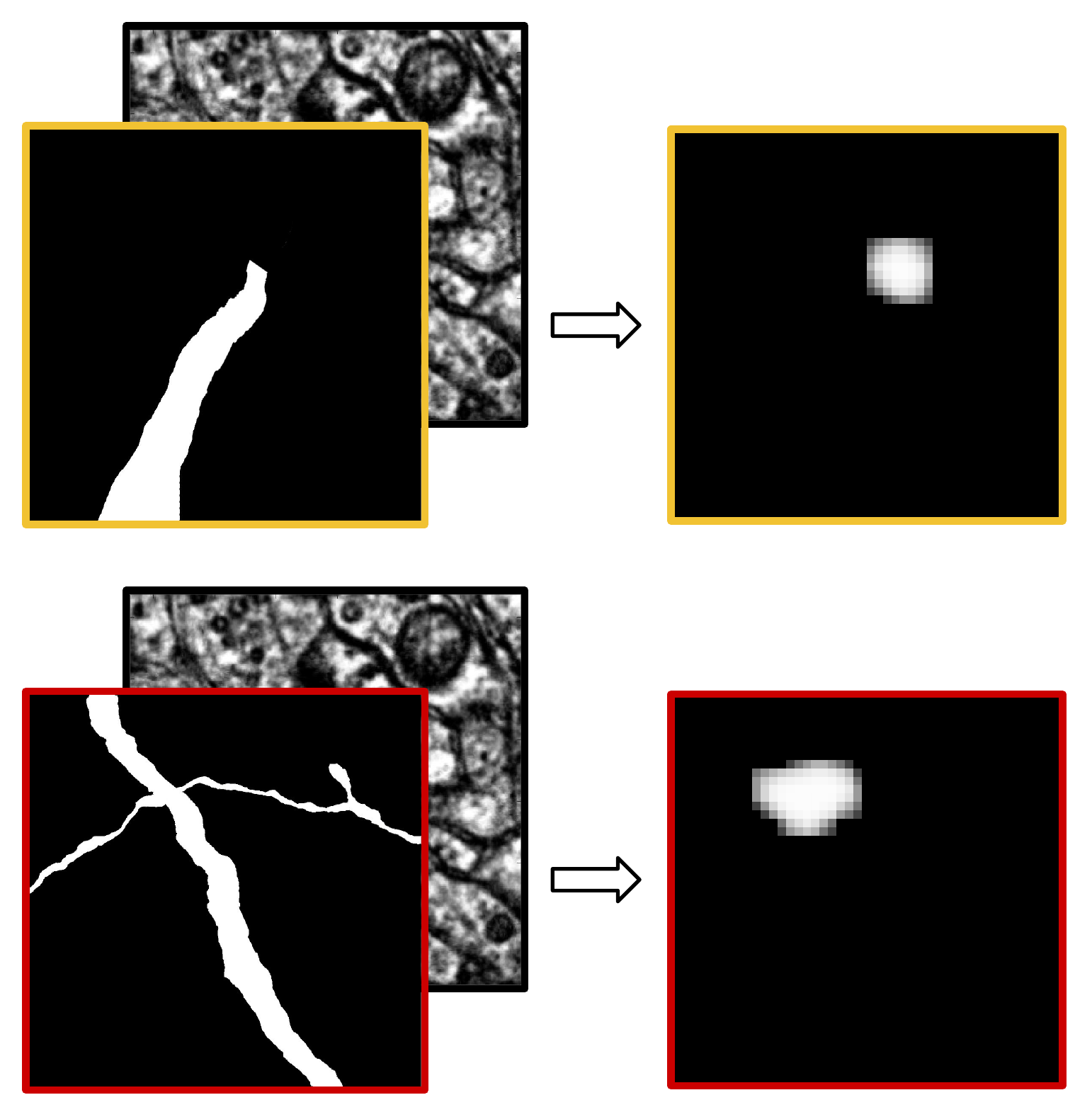}
		\caption{Error detection task for split (top) and merge (bottom) errors. The desired output is an error map. A voxel in the error map is lit up if and only if a window centered on it contains a split or merge error. We also consider a variant of the task in which the object mask is the sole input.}
		\label{fig:error_detection_cartoon}
	\end{subfigure}
\hfill
	\begin{subfigure}[t]{0.45\textwidth}
	\includegraphics[width=1.0\linewidth]{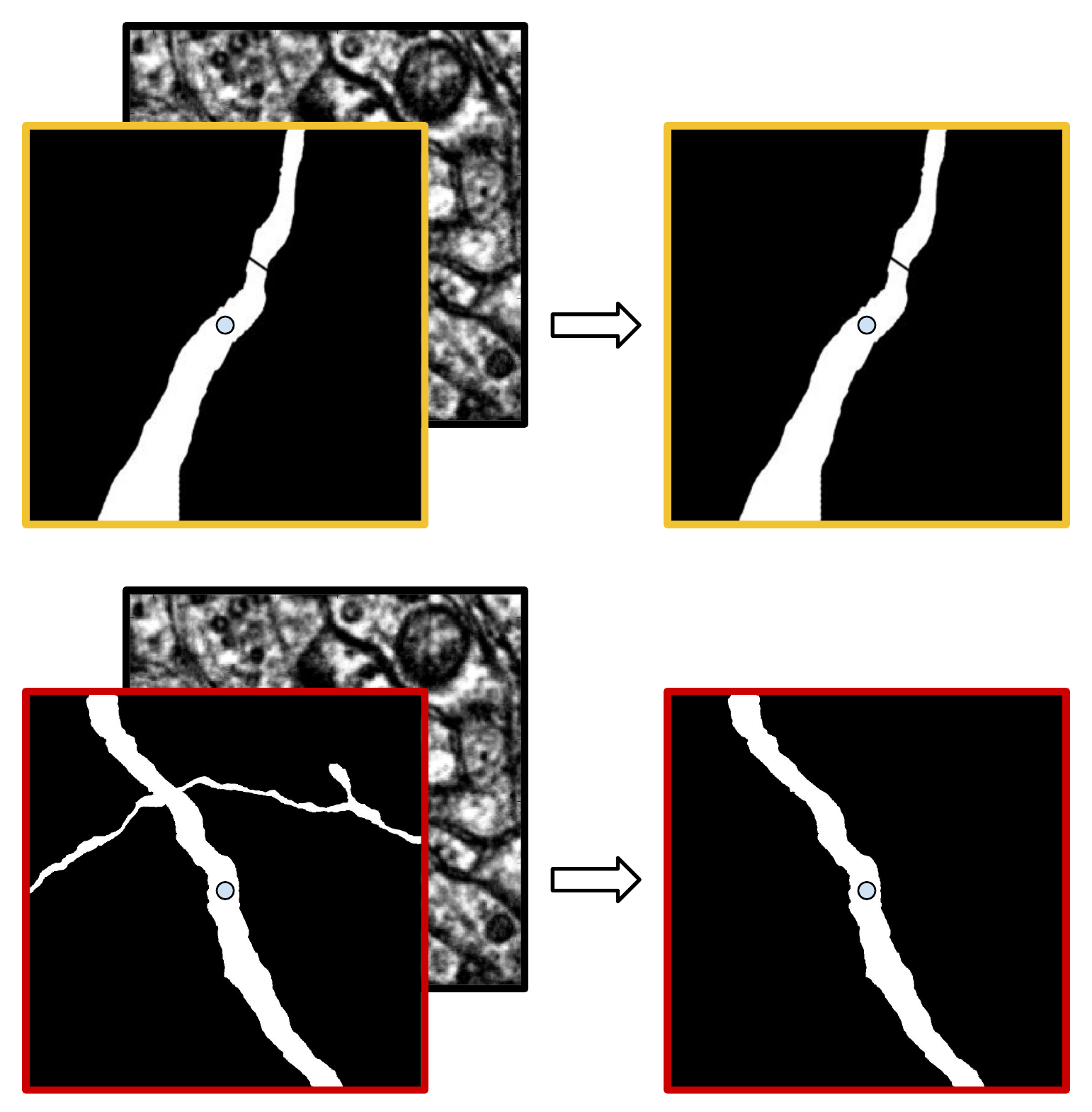}
	\caption{The object mask pruning task. The input mask is assumed to be a superset of a true object. The desired output (right) is the true object containing the central voxel (center circle). In the first case there is nothing to prune, while in the second case the object not overlapping the central voxel is erased.}
	\label{fig:error_correction_cartoon}
\end{subfigure}
        \end{center}
        \caption{Error detection and correction tasks.  For both tasks, the inputs are a candidate object mask and the original image. Note that diagrams are 2D for illustrative purposes, but in reality the inputs and outputs are 3D.}
\end{figure}

\section{Introduction}

While neuronal circuits can be reconstructed from volumetric electron
microscopic imagery, the process has historically \cite{white1986structure} and
even recently \cite{schmidt2017axonal} been highly laborious. One of the most
time-consuming reconstruction tasks is the tracing of the brain's ``wires,'' or
neuronal branches. This task is an example of instance segmentation, and can be
automated through computer detection of the boundaries between neurons.
Convolutional networks were first applied to neuronal boundary detection a
decade ago \cite{jain2007supervised, boundary_detection}. Since then
convolutional nets have become the standard approach, and the accuracy of
boundary detection has become impressively high  \cite{zeng2017deep,
beier2017multicut, kisuk, funke2017deep}.

Given the low error rates, it becomes helpful to think of subsequent processing steps in terms of modules that detect and correct errors. In the error detection task (Figure~\ref{fig:error_detection_cartoon}), the input is the raw image and a binary mask that represents a candidate object. The desired output is a map containing the locations of split and merge errors in the candidate object. Related work on this problem has been restricted to detection of merge errors only by either hand-designed \cite{multipass} or learned \cite{mergenet} computations. However, a typical segmentation contains both split and merge errors, so it would be desirable to include both in the error detection task.

In the error correction task (Figure~\ref{fig:error_correction_cartoon}), the input is again the raw image and a binary mask that represents a candidate object. The candidate object mask is assumed to be a superset of a true object, which is the desired output. With this assumption, error correction is formulated as \emph{object mask pruning}.
Object mask pruning can be regarded as the splitting of undersegmented objects to create true objects. In this sense, it is the opposite of agglomeration, which merges oversegmented objects to create true objects \cite{lash,gala}. Object mask pruning can also be viewed as the \emph{subtraction} of voxels from an object to create a true object. In this sense, it is the opposite of a flood-filling net \cite{floodfilling,januszewski2017high} or \emph{MaskExtend} \cite{multipass}, each iteration of which is the \emph{addition} of voxels to an object to create a true object. Iterative mask extension has been studied in other work on instance segmentation in computer vision \cite{recurrent_instance_seg_1, recurrent_instance_seg_2}.  The task of generating an object mask \emph{de novo} from an image has also been studied in computer vision~\cite{pinheiro2015}.

We implement both error detection and error correction using 3D multiscale convolutional networks.  One can imagine multiple uses for these nets in a connectomics pipeline. For example, the error-detecting net could be used to reduce the amount of labor required for proofreading by directing human attention to locations in the image where errors are likely. This labor reduction could be substantial because the declining error rate of automated segmentation has made it more time-consuming for a human to find an error.

We show that the error-detecting net can provide ``advice'' to the  error-correcting net in the following way. To create the candidate object mask for the error-correcting net from a baseline segmentation, one can simply take the union of all erroneous segments as found by the error-detecting net. Since the error rate in the baseline segmentation is already low, this union is small and it is easy to select out a single object. The idea of using the error detector to choose locations for the error corrector was proposed previously though not actually implemented \cite{multipass}. Furthermore, the idea of using the error detector to not only choose locations but provide ``advice'' is novel as far as we know.

We contend that our approach decomposes the neuron segmentation problem into two strictly easier pieces. First, we hypothesize that recognizing an error is much easier than producing the correct answer. Indeed, humans are often able to detect errors using only morphological cues such as abrupt terminations of axons, but may have difficulty actually finding the correct extension.

On the other hand, if the error-detecting net has high accuracy and the initial set of errors is sparse, then the error correction module only needs to prune away a small number of irrelevant parts from the candidate mask described above. This contrasts with the flood-filling task which involves an unconstrained search for new parts to add. Given that most voxels are \textit{not} a part of the object to be reconstructed, an upper bound on the object is usually more informative than a lower bound. As an added benefit, selective application of the error correction module near likely errors makes efficient use of our computational budget~\cite{multipass}.

In this paper, we support the intuition above by demonstrating high accuracy detection of both split and merge errors. We also demonstrate a complete implementation of the stated error detection-correction framework, and report significant improvements upon our baseline segmentation.

%Furthermore, we observe the interesting phenomenon that it is often much easier to detect an error than to find the correct segmentation. Indeed, humans are usually able to detect a segmentation error without even looking at the original image; they look for neurites that terminate prematurely or x-shaped junctions indicating incorrectly merged segments. However, the incredible density of information in neural tissue makes searching for the correction difficult.

Some of the design choices we made in our neural networks may be of interest to
other researchers. Our error-correcting net is trained to produce a vector field
via metric learning instead of directly producing an object mask. The vector
field resembles a semantic labeling of the image, so this approach blurs the
distinction between instance and semantic segmentation. This idea is relatively
new in computer vision
\cite{harley2015metric,fathi2017metric,brabandere2017metric}. Our multiscale
convolutional net architecture, while similar in spirit to the popular
U-Net~\cite{unet}, has some novelty. With proper weight sharing, our model can
be viewed as a feedback recurrent convolutional net unrolled in time (see the
appendix for details). Although our model architecture is closely related to the
independent works of~\cite{neuralfabric,mdnet,gridnet}, we contribute a
\emph{feedback recurrent} convolutional net interpretation.

\section{Error detection}
\subsection{Task specification: detecting split and merge errors}
\label{sec:detection_spec}
Given a single segment in a proposed segmentation presented as an object mask $Obj$, the error detection task is to produce a binary image called the \textit{error map}, denoted $Err_{p_x\times p_y \times p_z}(Obj)$. The definition of the error map depends on a choice of a window size $p_x \times p_y \times p_z$. A voxel $i$ in the error map is 0 if and only if the restriction of the input mask to a window centred at $i$ of size $p_x \times p_y \times p_z$ is voxel-wise equal to the restriction of some object in the ground truth. Observe that the error map is sensitive to both split and merge errors.

A smaller window size allows us to localize errors more precisely. On the other hand, if the window radius is less than the width of a typical boundary between objects, it is possible that two objects participating in a merge error never appear in the same window. These merge errors would not be classified as an error in any window.

We could use a less stringent measure than voxel-wise equality that disregards small perturbations of the boundaries of objects. However, our proposed segmentations are all composed of the same building blocks (supervoxels) as the ground truth segmentation, so this is not an issue for us.

We define the \textit{combined error map} as $\sum_{Obj} Err(Obj) * Obj$ where $*$ represents pointwise multiplication. In other words, we restrict the error map for each object to the object itself, and then sum the results. The figures in this paper show the \textit{combined error map}.

\subsection{Architecture of the error-detecting net}
\begin{figure}
\centering
\includegraphics[width=1.0\linewidth]{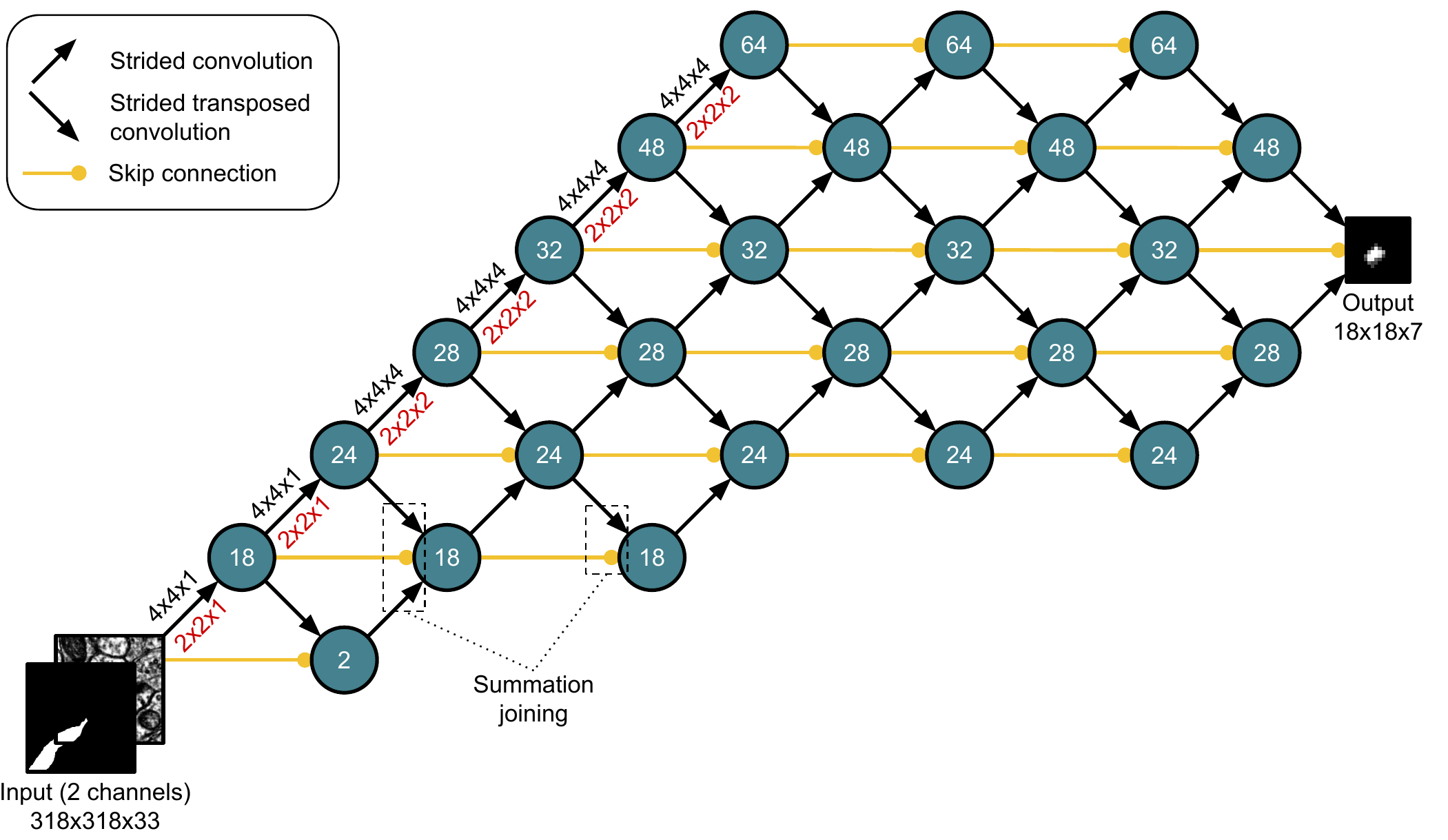}
\centering
\includegraphics[width=1.0\linewidth]{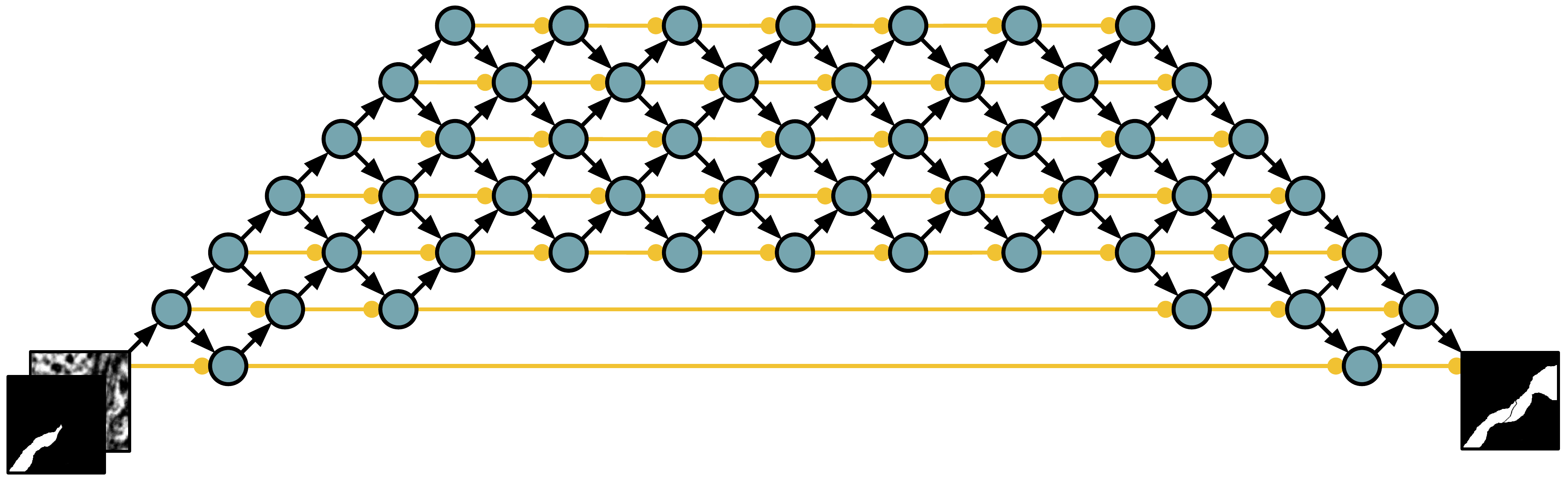}

\caption{Architectures for the error-detecting and error-correcting nets
respectively. Each node represents a layer and the number inside represents the
number of feature maps. The layers closer to the top of the diagram have lower
resolution than the layers near the bottom. We make savings in computation by
minimizing the number of high resolution feature maps. The diagonal arrows
represent strided convolutions, while the horizontal arrows represent skip
connections. Associated with the diagonal arrows, black numbers indicate filter
size and red numbers indicate strides in $x$, $y$ and $z$ dimension. Due to the
anisotropy of the resolution of the images in our dataset, we design our nets so
that the first convolutions are exclusively 2D while later convolutions are 3D.
The field of view of a unit in the higher layers is therefore roughly cubic. To
limit the number of parameters in our model, we factorize all 3D convolutions
into a 2D convolution followed by a 1D convolution in $z$-dimension. We also use
weight sharing between some convolutions at the same height. Note that the
error-correcting net is a prolonged, symmetric version of the error-detecting
net. For more detail of the error corrector, see the appendix.}

\label{fig:architecture}
\end{figure}

We take a fully supervised approach to error detection. We implement error
detection using a multiscale 3D convolutional network. The architecture is
detailed in Figure~\ref{fig:architecture}. Its design is informed by experience
with convolutional nets for neuronal boundary detection~\cite{kisuk} and
reflects recent trends in neural network design \cite{unet,resnet}. Its field of
view is $P_x\times P_y\times P_z=318\times 318\times 33$ (which is roughly cubic
in physical size given the anisotropic resolution of our dataset). The network
computes (a downsampling of) $Err_{46 \times 46 \times 7}$. At test time, we
perform inference in overlapping windows and conservatively blend the output
from overlapping windows using a maximum operation.

We trained two variants, one of which takes as input only $Obj$, and another which additionally receives as input the raw image.

\section{Error correction}
\subsection{Task specification: object mask pruning}
Given an image patch of size $P_x\times P_y\times P_z$ and a candidate object mask of the same dimensions, the $\textit{object mask pruning}$ task is to erase all voxels which do not belong to the true object overlapping the central voxel.  The candidate object mask is assumed to be a superset of the true object.

\subsection{Architecture of the error-correcting net}
Yet again, we implement error correction using a multiscale 3D convolutional network. The architecture is detailed in Figure~\ref{fig:architecture}. One difficulty with training a neural network to reconstruct the object containing the central voxel is that the desired output can change drastically as the central voxel moves between objects. We use an intermediate representation whose role is to soften this dependence on the location of the central voxel. The desired intermediate representation is a $k=6$ dimensional vector $v(x,y,z)$ at each point $(x,y,z)$ such that points within the same object have similar vectors and points in different objects have different vectors. We transform this vector field into a binary image $M$ representing the object overlapping the central voxel as follows:
\begin{equation*}
	M(x,y,z)=\exp\left( -||v(x,y,z)-v(0,0,0)||^2 \right),
\end{equation*}
where $(0,0,0)$ is the central voxel. When an over-segmentation is available, we replace $v(0,0,0)$ with the average of $v$ over the supervoxel containing the central voxel. This trick makes it unnecessary to centre our windows far away from a boundary, as was necessary in \cite{floodfilling}. Note that we backpropagate through the transform $M$, so the vector representation may be seen as an implementation detail and the final output of the network is just a (soft) binary image.

\section{How the error detector can ``advise'' the error corrector}
\begin{figure}
\begin{center}
	\includegraphics[width=0.85\linewidth]{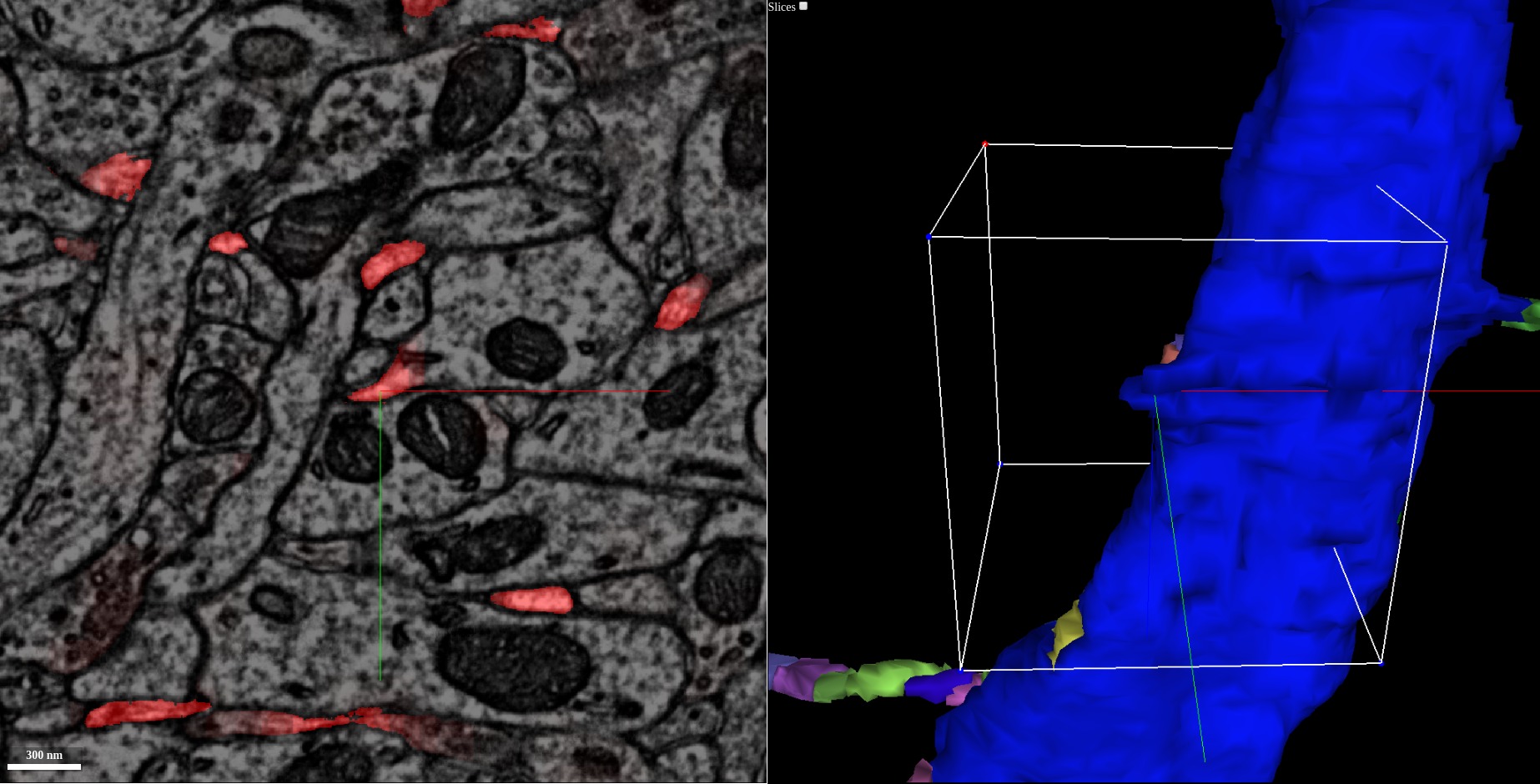}
	\caption{An example of a mistake in the initial segmentation. The dendrite is missing a spine. The red overlay on the left shows the combined error map (defined in Section~\ref{sec:detection_spec}); the stump in the centre of the image was clearly marked as an error.}

	\includegraphics[width=0.85\linewidth]{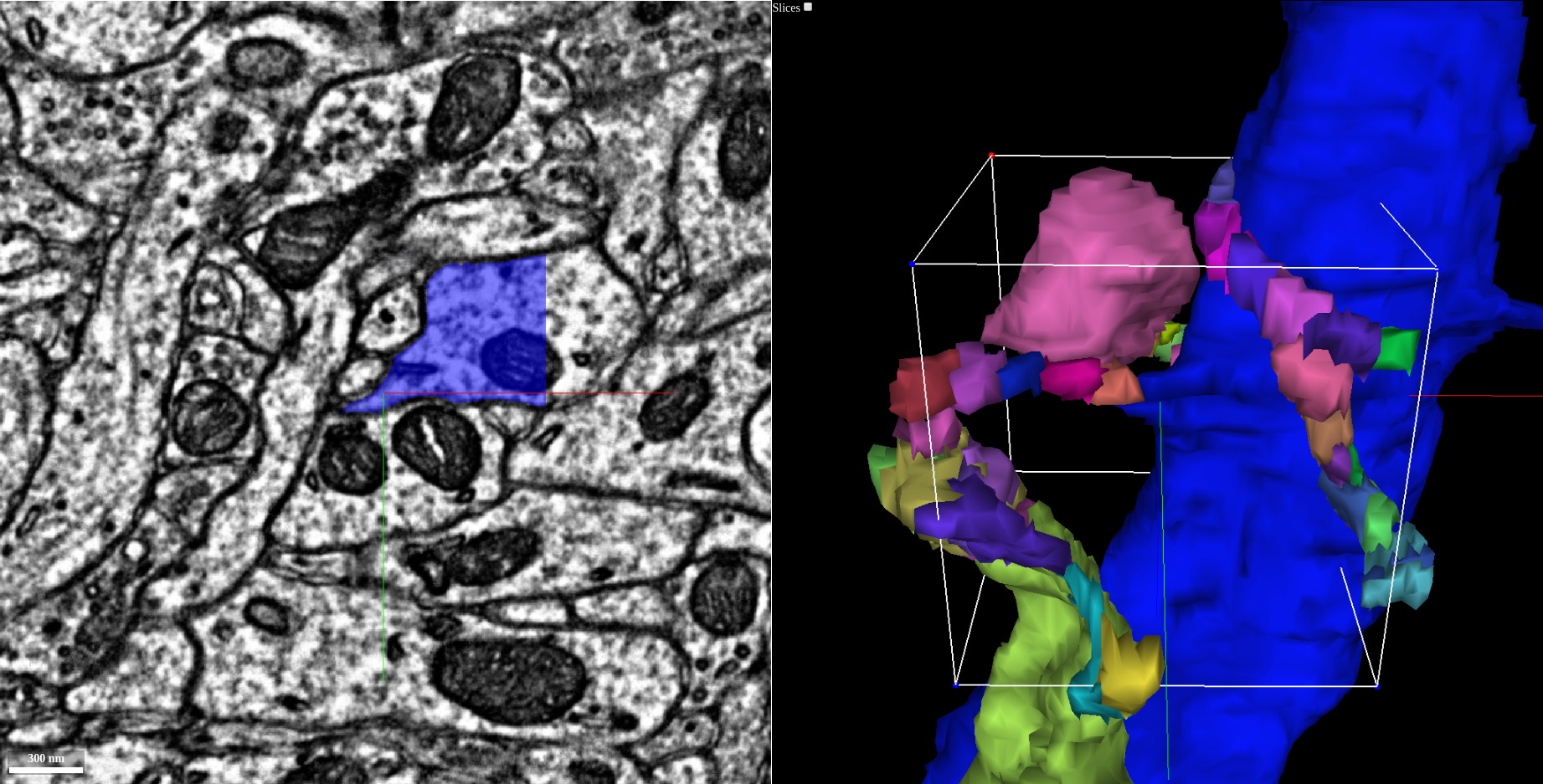}
	\caption{The right shows all objects which contained  a detected error in the vicinity. For clarity, each supervoxel was drawn with a different colour. The union of these objects is the binary mask which is provided as input to the error correction network. For clarity, these objects were clipped to lie within the white box representing the field of view of our error correction network. The output of the error correction network is overlaid in blue on the left.}

	\includegraphics[width=0.85\linewidth]{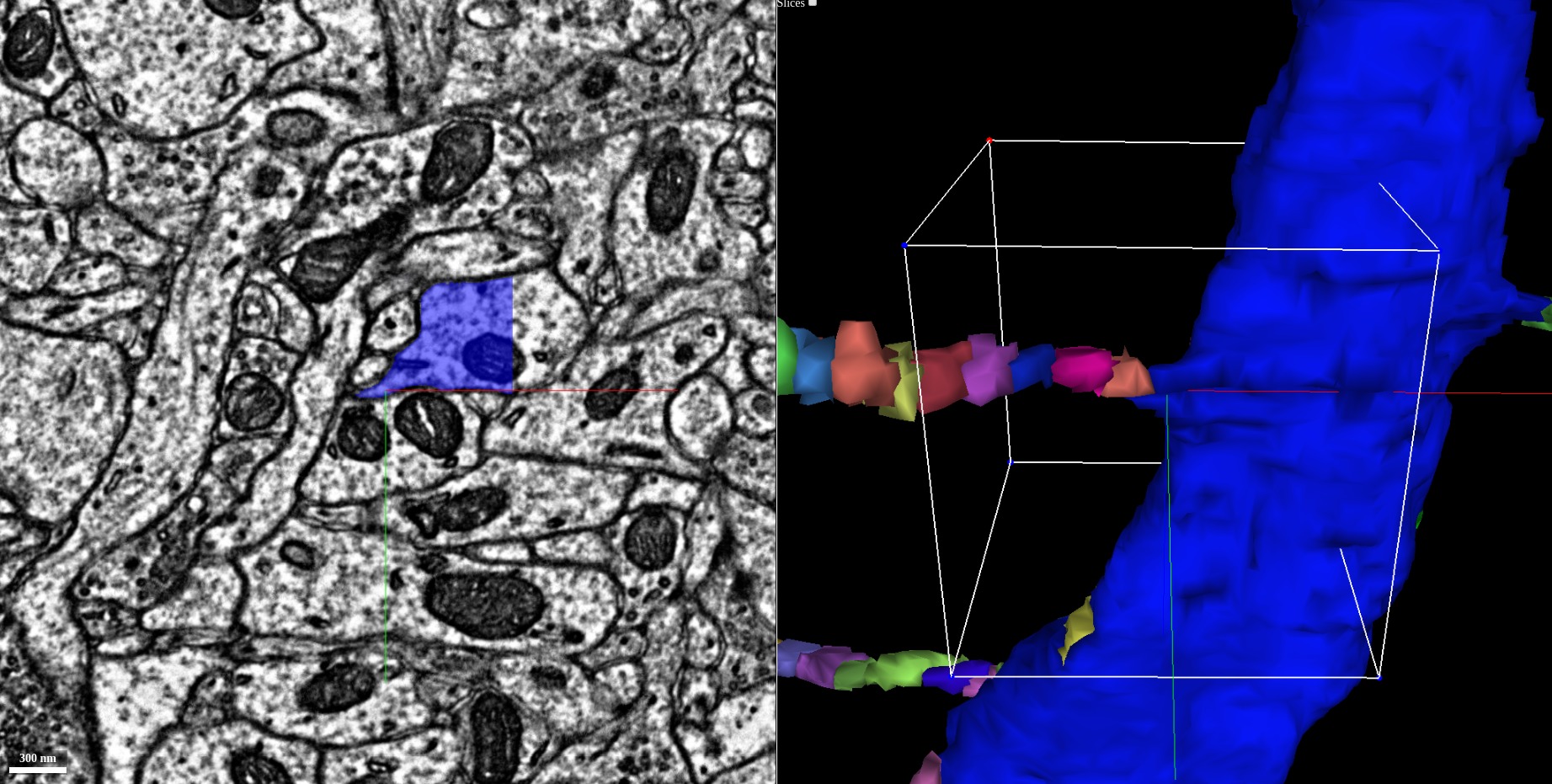}
	\caption{The supervoxels assembled in accordance with the output of the error correction network.}
\end{center}
\end{figure}

Suppose that we would like to correct the errors in a baseline segmentation.
Obviously, the error-detecting net can be used to find locations where the
error-correcting net can be applied \cite{multipass}.  Less obviously, the
error-detecting net can be used to construct the object mask that is the input
to the error-correcting net. We refer to this object mask as the ``advice mask''
and its construction is important because the baseline object to be corrected
might contain split as well as merge errors, while the object mask pruning task
can correct only merge errors.

The advice mask is defined as the union of the baseline object at the central
pixel with all other baseline objects in the window that contain errors as
judged by the error-detecting net. The advice mask is a superset of the true
object overlapping the central voxel, assuming that the error-detecting net
makes no mistakes.  Therefore advice is suitable as an input to the object mask
pruning task.

The details of the above procedure are as follows. We begin with an initial baseline segmentation whose remaining errors are assumed to be sparsely distributed. During the error correction phase, we iteratively update a segmentation represented as the connected components of a graph $G$ whose vertices are segments in a strict over-segmentation (henceforth called supervoxels). We also maintain the combined error map associated with the current segmentation. We binarize the error map by thresholding it at 0.25.

Now we iteratively choose a location $\ell=(x,y,z)$ which has value 1 in the binarized combined error map. In a $P_x \times P_y \times P_z$ window centred on $\ell$, we prepare an input for the error corrector by taking the union of all segments containing at least one white voxel in the error map. The error correction network produces from this input a binary image $M$ representing the object containing the central voxel. For each supervoxel $S$ touching the central $P_x/2 \times P_y/2 \times P_z/2$ window, let $M(S)$ denote the average value of $M$ inside $S$. If $M(S) \not \in [0.1,0.9]$ for all $S$ in the relevant window (i.e. the error corrector is confident in its prediction for each supervoxel), we add to $G$ a clique on $\{S \mid M(S) > 0.9\}$ and delete from $G$ all edges between $\{S \mid M(S) < 0.1\}$ and $\{S \mid M(S) > 0.9\}$. The effect of these updates is to change $G$ to locally agree with $M$. Finally, we update the combined error map by applying the error detector at all locations where its decision could have changed.

We iterate until every location is zero in the error map or has been covered by a window at least $t=2$ times by the error corrector. This stopping criterion guarantees that the algorithm terminates. In practice, the segmentation converges without this auxiliary stopping condition to a state in which the error corrector fails confidence threshold everywhere. However, it is hard to certify convergence since it is possible that the error corrector could give different outputs on slightly shifted windows. Based on our validation set, increasing $t$ beyond 2 did not measurably improve performance.

Note that this algorithm deals with split and merge errors, but cannot fix errors already present at the supervoxel level.

%In this section, we present a greedy algorithm which combines the error detector and the error corrector to greedily update an initial segmentation. We assume that we are provided with an initial segmentation along with a strict over-segmentation. We term the segments in the over-segmentation supervoxels. Throughout the error correction phase, we maintain a graph whose vertices are supervoxels and whose connected components are segments in the proposed segmentation.

%Let $L$ be a list of locations densely sampled from the image. We say that an error is detected at a location of the the error detector reports a value of >0.25 on a window centred at that location. For each location in the

\section{Experiments}
\subsection{Dataset}

%Our dataset was culled from an unfortunate soul named Pinky. He had a big heart but a small brain, which made him perfect for our experiment. May he rest in peace.

Our dataset is a sample of mouse primary visual cortex (V1) acquired using
serial section transmission electron microscopy (TEM) at the Allen Institute for
Brain Science. The voxel resolution is $3.6~\text{nm} \times 3.6~\text{nm}
\times 40~\text{nm}$.

Human experts used the VAST software tool \cite{kasthuri2015saturated, VAST} to densely reconstruct multiple volumes that amounted to $530$ Mvoxels of ground truth annotation. These volumes were used to train a neuronal boundary detection network (see the appendix for architecture). We applied the resulting boundary detector to a larger volume of size $4800$ Mvoxels to produce a preliminary segmentation, which was then proofread by the tracers. This bootstrapped ground truth was used to train the error detector and corrector. A subvolume of size $910$ Mvoxels was reserved for validation, and a subvolume of size $910$ Mvoxels was reserved for testing.

Producing the gold standard segmentation required a total of $\sim 560$ tracer hours, while producing the bootstrapped ground truth required $\sim 670$ tracer hours.

\subsection{Baseline segmentation}
Our baseline segmentation was produced using a pipeline of multiscale
convolutional networks for neuronal boundary detection, watershed, and mean
affinity agglomeration~\cite{kisuk}. We describe the pipeline in detail in the
appendix. The segmentation performance values reported for the baseline are
taken at a mean affinity agglomeration threshold of 0.23, which minimizes the
\textsl{variation of information} error metric~\cite{meila2007,vi} on the test
volumes.

\subsection{Training procedures}
\paragraph{Sampling procedure}
\label{sec:sampling}
Here we describe our procedure for choosing a random point location in a
segmentation. Uniformly random sampling is unsatisfactory since large objects
such as dendritic shafts will be overrepresented. Instead, given a segmentation,
we sample a location $(x,y,z)$ with probability inversely proportional to the
fraction of a window of size $128 \times 128 \times 16$ centred at $(x,y,z)$
which is occupied by the object containing the central
voxel~\cite{floodfilling}.

\paragraph{Training of error detector} An initial segmentation containing errors
was produced using our baseline neuronal boundary detector combined with mean
affinity agglomeration at a threshold of 0.3. Point locations were sampled
according to the sampling procedure specified above. We augmented all of our
data with rotations and reflections. We used a pixelwise cross-entropy loss.

\paragraph{Training of error corrector} We sampled locations in the ground truth
segmentation as described above. At each location $\ell = (x,y,z)$, we generated
a training example as follows. Let $Obj_\ell$ be the ground truth object
touching $\ell$. We selected a random subset of the objects in the window
centred on $\ell$ including $Obj_\ell$. To be specific, we chose a number $p$
uniformly at random from $[0,1]$, and then selected each segment in the window
with probability $p$ in addition $Obj_\ell$. The input at $\ell$ was then a
binary mask representing the union of the selected objects along with the raw EM
image, and the desired output was a binary mask representing only $Obj_\ell$.
The dataset was augmented with rotations, reflections, simulated misalignments
and missing sections~\cite{kisuk}. We used a pixelwise cross-entropy loss.

Note that this training procedure uses only the ground truth segmentation and is completely independent of the error detector and the baseline segmentation. This convenient property is justified by the fact that if the error detector is perfect, the error corrector only ever receives as input unions of complete objects.

\subsection{Error detection results}
\label{sec:detection_results}

\begin{figure}
\begin{center}
	\includegraphics[width=0.8\linewidth]{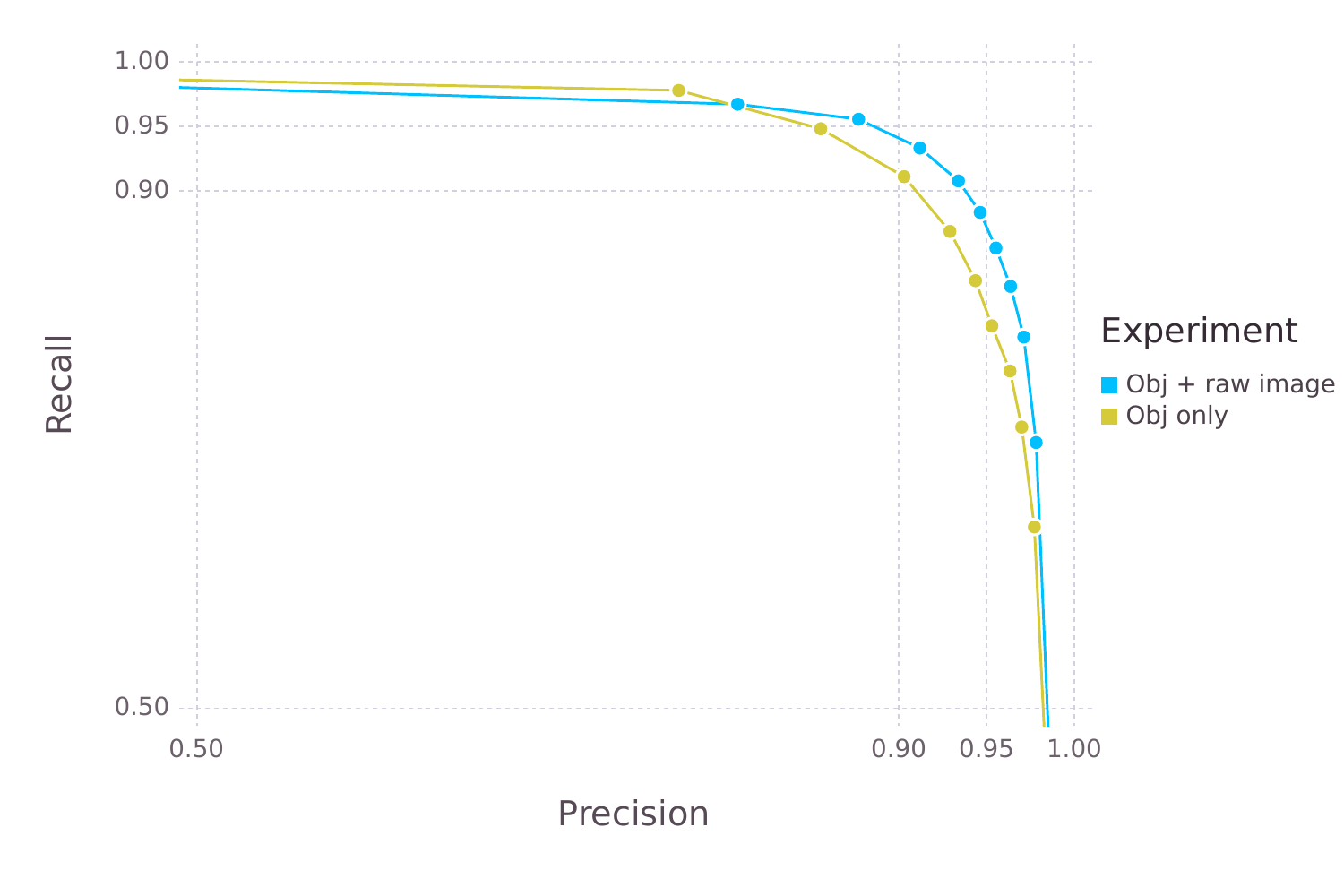}
	\caption{Precision and recall for error detection, both with and without access to the raw image. In the test volume, there are $8248$ error free locations and $944$ locations with errors. In practice, we use threshold which guarantees >~95\% recall and >~85\% precision.}
	\label{fig:error_detection_pr}
\end{center}
\end{figure}

To measure the quality of error detection, we densely sampled points in our test
volume as in Section~\ref{sec:sampling}. In order to remove ambiguity over the
precise location of errors, we filtered out points which contained an error
within a surrounding window of size $80\times80\times 8$ but not a window of
size $40\times 40 \times 4$. These locations were all unique, in that two
locations in the same object were separated by at least $80,80,8$ in $x,y,z$,
respectively. Precision and recall simultaneously exceed 90\% (Figure
\ref{fig:error_detection_pr}). Empirically, many of the false positive examples
come from where a dendritic spine head curls back and touches its trunk. These
examples locally appear to be incorrectly merged objects.

We trained one error detector with access to the raw image and one without. The
network's admirable performance even without access to the image as seen in
Figure~$\ref{fig:error_detection_pr}$ supports our hypothesis that error
detection is a relatively easy task and can be performed using only shape cues.

\begin{figure}
	\begin{minipage}[t]{0.47\textwidth}
\begin{center}
	\includegraphics[width=1.0\linewidth]{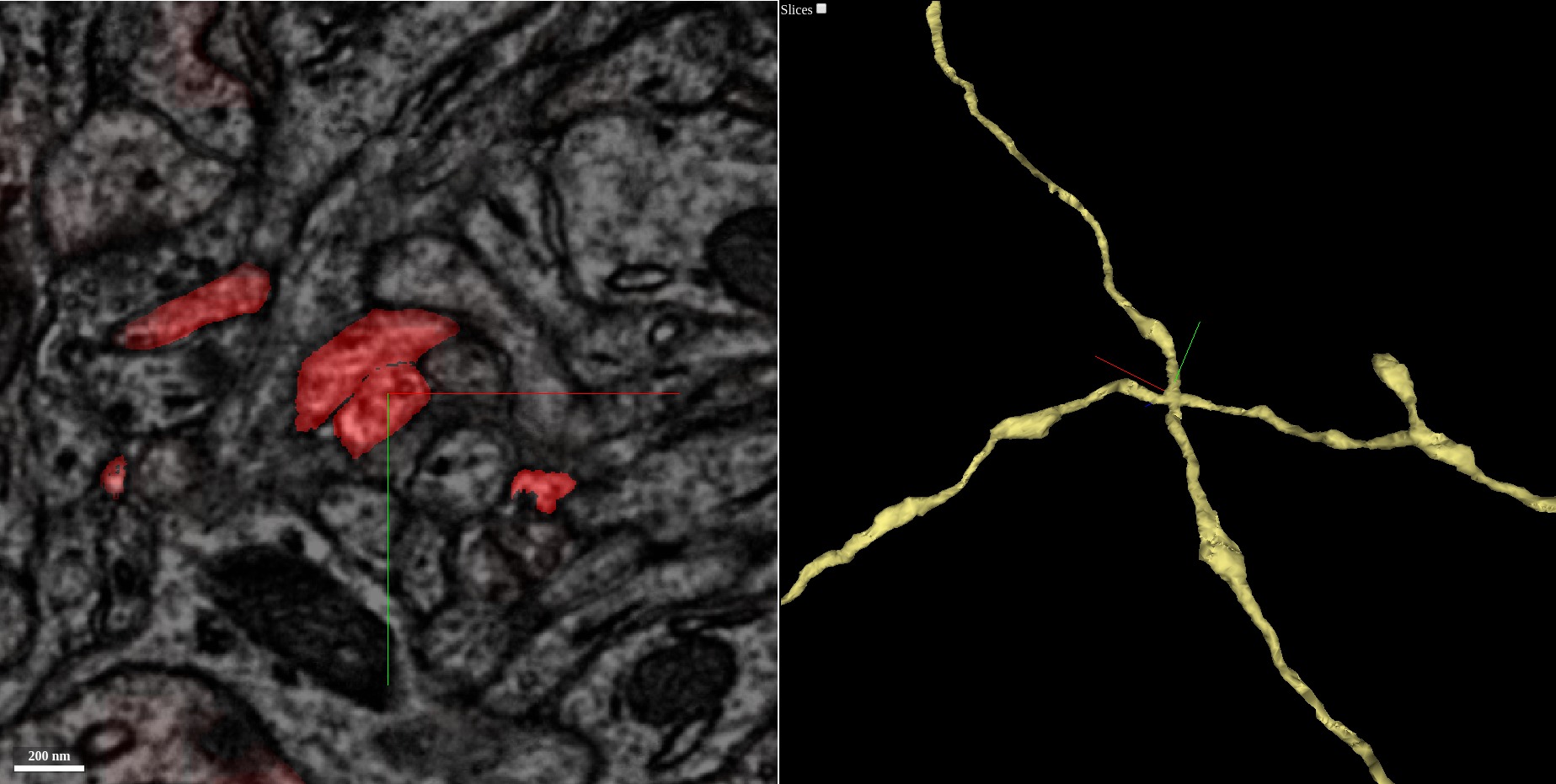}
	\caption{An example of a detected error. The right shows two incorrectly
	merged axons, and the left shows the predicted combined error map (defined
	in Section~\ref{sec:detection_spec}) overlaid on the corresponding 2D image
	in red.}
    \label{fig:x_error}
\end{center}
\end{minipage}
\hfill
	\begin{minipage}[t]{0.47\textwidth}
\begin{center}
	\includegraphics[width=1.0\linewidth]{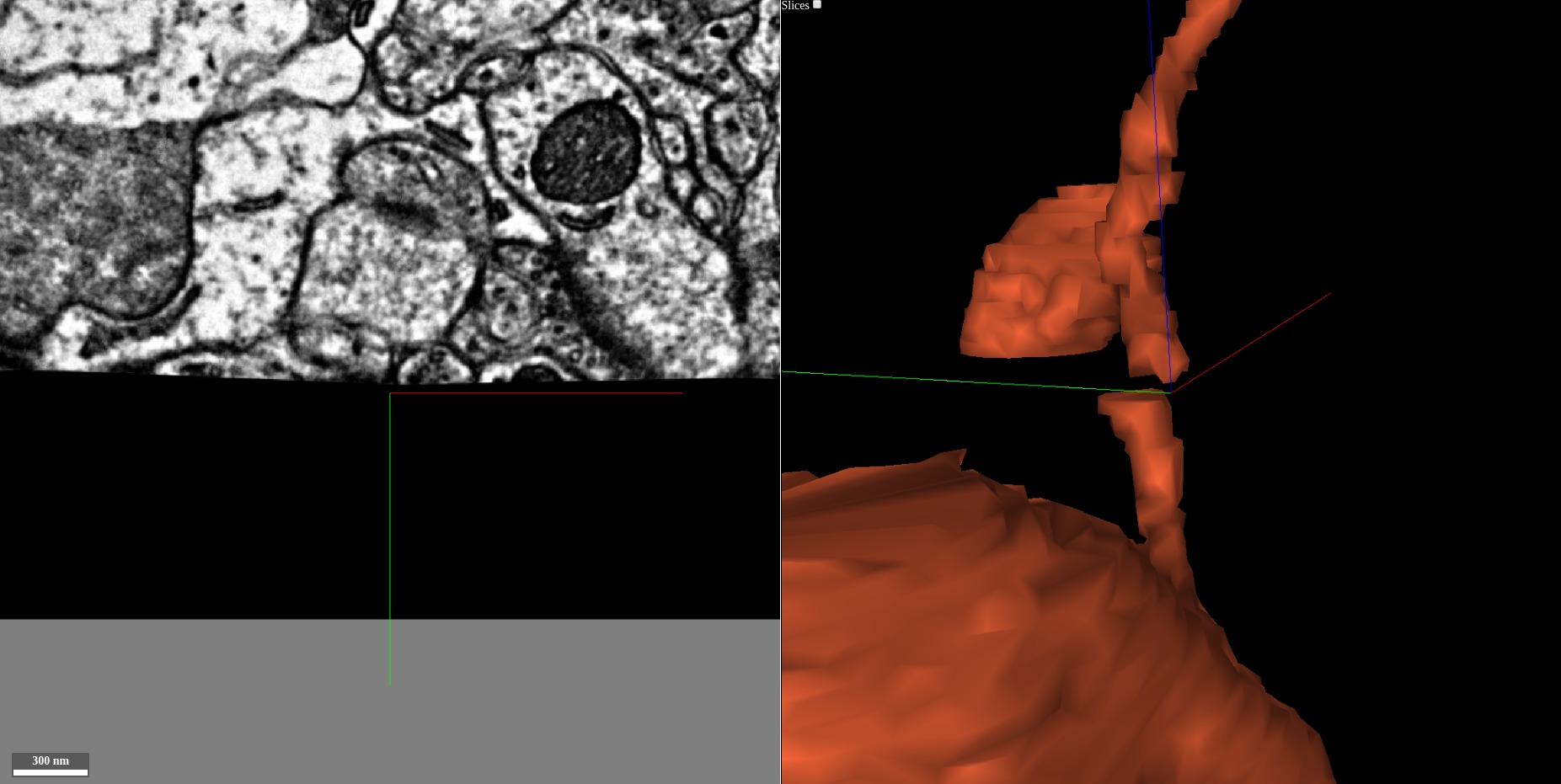}
	\caption{A difficult location with missing data in one section combined with a misalignment between sections. The error-correcting net was able to trace across the missing data.}
	\label{fig:difficult}
\end{center}
\end{minipage}
\end{figure}

Merge errors qualitatively appear to be especially easy for the network to detect; an example is shown in Figure~\ref{fig:x_error}.

\subsection{Error correction results}
\begin{table}[!b]
  \caption{Comparing segmentation performance}
  \label{table:vi_scores}
  \centering
  \begin{tabular}{lllll}
    \toprule
	& $VI_\text{merge}$ & $VI_\text{split}$ & Rand Recall & Rand Precision\\
    \midrule
    Baseline & 0.162 & 0.142 & 0.952 & 0.954\\
    Without Advice & 0.130 & 0.057 & 0.956 & 0.979\\
	With Advice & \textbf{0.088} & \textbf{0.052} & \textbf{0.974} & \textbf{0.980}\\
    \bottomrule
  \end{tabular}
\end{table}

In order to demonstrate the importance of error detection to error correction,
we ran two experiments: one in which the binary mask input to the error
corrector was simply the union of all segments in the window (``without
advice''), and one in which the binary mask was the union of all segments with a
detected error (``with advice''). In the ``without advice'' mode, the network is
essentially asked to reconstruct the object overlapping the central voxel in one
shot. Table \ref{table:vi_scores} shows that advice confers a considerable
advantage in performance on the error corrector.

\begin{figure}
\begin{center}
\includegraphics[width=0.85\linewidth]{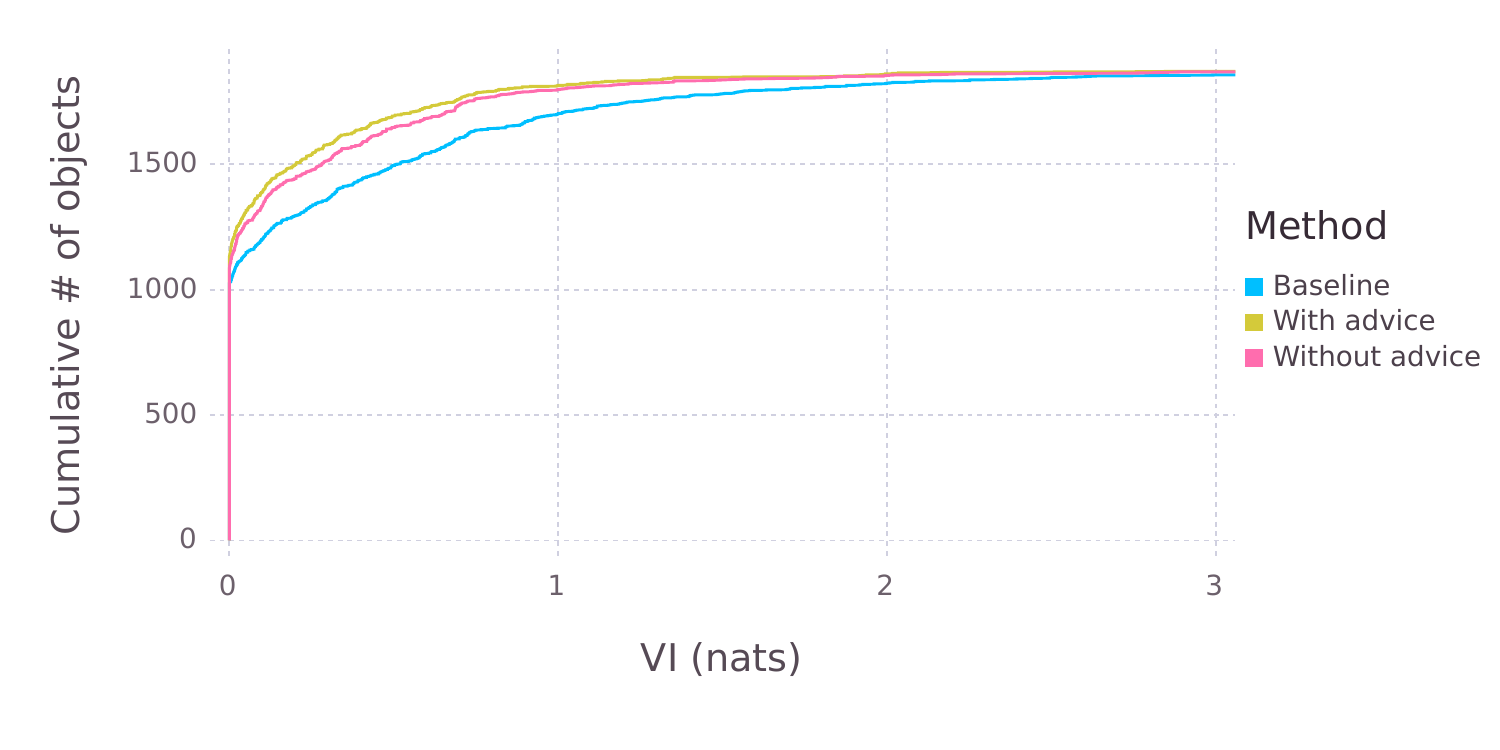}
\caption{Per-object VI scores for the 940 reconstructed objects in our test volume. Almost 800 objects are completely error free in our segmentation. These objects are likely all axons; almost every dendrite is missing a few spines.}
\label{fig:decomp_vi_scores}
\end{center}
\end{figure}

It is sometimes difficult to assess the significance of an improvement in the
variation of information or \textsl{Rand score}~\cite{rand1971,ignacio2015}
since changes can be dominated by modifications to a few large objects.
Therefore, we decomposed the variation of information into a score for each
object in the ground truth. Figure \ref{fig:decomp_vi_scores} summarizes the
cumulative distribution of the values of
$VI(i)=VI_\text{merge}(i)+VI_\text{split}(i)$ for all segments $i$ in the ground
truth. See the appendix for a precise definition of $VI(i)$.

The number of errors from the set in Section~$\ref{sec:detection_results}$ that
were fixed or introduced by our iterative refinement procedure is shown in
Figure~\ref{table:errors_fixed}. These numbers should be taken with a grain of
salt since topologically insignificant changes could result in errors.
Regardless, it is clear that our iterative refinement procedure fixed a
significant fraction of the remaining errors and that ``advice'' improves the
error corrector.

The results are qualitatively impressive as well. The error-correcting network
is sometimes able to correctly merge disconnected objects, as exemplified in
Figure~\ref{fig:difficult}.

\begin{table}[h]
  \caption{Number of errors fixed and introduced relative to the baseline}
  \label{table:errors_fixed}
  \centering
  \begin{tabular}{lccc}
    \toprule
	& \# Errors & \# Errors fixed & \# Errors introduced\\
    \midrule
    Baseline & 944 & - & - \\
    Without Advice & 474 & 547 & 77\\
	With Advice & \textbf{305} & 707 & 68\\
    \bottomrule
  \end{tabular}
\end{table}

%Given that our baseline approach already produces state-of-the-art results on other datasets (see \cite{kisuk}), we expect that the method presented here is a substantial improvement upon the state of the art. However, we have not conducted experiments on publicly available datasets, and therefore we leave a careful comparison for future work.

\begin{table}[!h]
	\caption{Computation time for a $2048\times 2048\times 256$ volume using a single TitanX Pascal GPU}
\label{table:timing}
  \centering
  \begin{tabular}{ccc}
    \toprule
	Boundary detection & Error detection & Error correction\\
	\midrule
	18 mins & 25 mins & 55 mins\\
	\bottomrule
  \end{tabular}
\end{table}

\subsection{Computational cost analysis}
Table \ref{table:timing} shows the computational cost of the most expensive
parts of our segmentation pipeline. Boundary detection and error detection are
run on the entire image, while error correction is run on roughly 10\% of the
possible locations in the image.  Error correction is still the most costly
step, but it would be $10\times$ more costly without restricting to the
locations found by the error-detecting network. Therefore, the cost of error
detection is more than justified by the subsequent savings during the error
correction phase.  The number of locations requiring error correction will fall
even further if the precision of the error detector increases or the error rate
of the initial segmentation decreases.

% \begin{table}[h]
% 	\caption{Computation time for a $2048\times 2048\times 256$ volume using a single TitanX Pascal GPU}
% \label{table:timing}
%   \centering
%   \begin{tabular}{ll}
% 	  \toprule
% 	Boundary Detection & 18 mins\\
% 	\midrule
% 	Error Detection & 25 mins\\
% 	\midrule
% 	Error Correction & 55 mins\\
% 	\bottomrule
%   \end{tabular}
% \end{table}

\section{Conclusion and future directions}
We have developed an error detector for the neuronal segmentation problem and
combined it with an error correction module. In particular, we have shown that
our error detectors are able to exploit priors on neuron shape, having
reasonable performance even without access to the raw image. We have made
significant savings in computation by applying expensive error correction
procedures only where predicted necessary by the error detector. Finally, we
have demonstrated that the ``advice'' of error detection improves an error
correction module, improving segmentation performance upon our baseline.

We expect that significant improvements in the accuracy of error detection could
come from aggressive data augmentation. We can mutilate a ground truth
segmentation in arbitrary (or even adversarial) ways to produce unlimited
examples of errors.

An error detection module has many potential uses beyond the ones presented
here. For example, we could use error detection to direct ground truth
annotation effort toward mistakes. If sufficiently accurate, it could also be
used directly as a learning signal for segmentation algorithms on unlabelled
data. The idea of co-training our error-detecting and error-correcting nets is
natural in view of recent work on generative adversarial networks
\cite{cgan1,cgan2}.

%Our approach may also be compared with models for visual attention in the literature (for example, \cite{recurrent_attention}). Recurrent neural networks are able to learn in an end-to-end way how to find which parts of an image are relevant to the given task. In contrast, we have a fixed policy for which objects to attend to: we attend to those objects which likely contain errors. One of our central findings is that this policy is highly selective and improves segmentation performance.

%We also argue that the error detection task is better posed than the supervoxel agglomeration task. Given two objects which already contain errors, it is often unclear whether the segmentation improves after they have been merged. In \cite{lash}, the authors resolve this ambiguity by training to predict the change in Rand score from a proposed merge.

%While $\cite{floodfilling}$ performs inference densely, we selectively apply our error correction module near likely errors. This comparatively reduces our computational cost. We sacrifice end-to-end training for this advantage.

%Our main novelty relative to their approach is the use of the advice of an error detector to bias flood-filling. While they present their network with a partially reconstructed object and ask for a completion, we present our network with the union of all possibly incorrect segments in a window and ask the network to split out a single object. Since the typical error rate is already low, this ``advice'' on which objects to consider is informative and significantly improves the performance of our flood-filling networks.
\subsubsection*{Author contributions \& acknowledgements}
JZ conceptualized the study and conducted most of the experiments and
evaluation. IT (along with Will Silversmith) created much of the infrastructure
necessary for visualization and running our algorithms at scale. KL produced the
baseline segmentation. HSS helped with the writing.

% \subsubsection*{Acknowledgements}
We are grateful to Clay Reid, Nuno da Costa, Agnes Bodor, Adam Bleckert, Dan
Bumbarger, Derrick Britain, JoAnn Buchannan, and Marc Takeno for acquiring the
TEM dataset at the Allen Institute for Brain Science. The ground truth
annotation was created by Ben Silverman, Merlin Moore, Sarah Morejohn, Selden
Koolman, Ryan Willie, Kyle Willie, and Harrison MacGowan. We thank Nico Kemnitz
for proofreading a draft of this paper. We thank Jeremy Maitin-Shepard at Google
and the other contributors to the Neuroglancer project for creating an
invaluable visualization tool.

We acknowledge NVIDIA Corporation for providing us with early access to Titan X
Pascal GPU used in this research, and Amazon for assistance through an AWS
Research Grant. This research was supported by the Mathers Foundation, the
Samsung Scholarship and the Intelligence Advanced Research Projects Activity
(IARPA) via Department of Interior/ Interior Business Center (DoI/IBC) contract
number D16PC0005. The U.S. Government is authorized to reproduce and distribute
reprints for Governmental purposes notwithstanding any copyright annotation
thereon. Disclaimer: The views and conclusions contained herein are those of the
authors and should not be interpreted as necessarily representing the official
policies or endorsements, either expressed or implied, of IARPA, DoI/IBC, or the
U.S. Government.

\bibliography{bib}

\section*{Supplementary Information for ``An Error Detection and Correction Framework for Connectomics'' by Zung et al. (\textsl{NIPS 2017)}}

% \begin{appendices}
\appendix
\renewcommand\thefigure{S\arabic{figure}}

%-------------------------------------------------------------------------------
% Section A. Baseline Neuronal Boundary Detection
%-------------------------------------------------------------------------------
\section{Baseline neuronal boundary detection}
\label{appendix:baseline}

In this section, we describe our baseline segmentation pipeline, which is
similar to what is described in~\cite{kisuk}. The major difference is our novel
densely multiscale 3D convolutional network architecture for neuronal boundary
detection, which is described below. (The same \emph{class} of architecture
was employed in error detection and error correction. See main text.)

\subsection{Network architecture}
\label{sec:deltanet}

\begin{figure}[!b]
\centering
\includegraphics[width=1.0\linewidth]{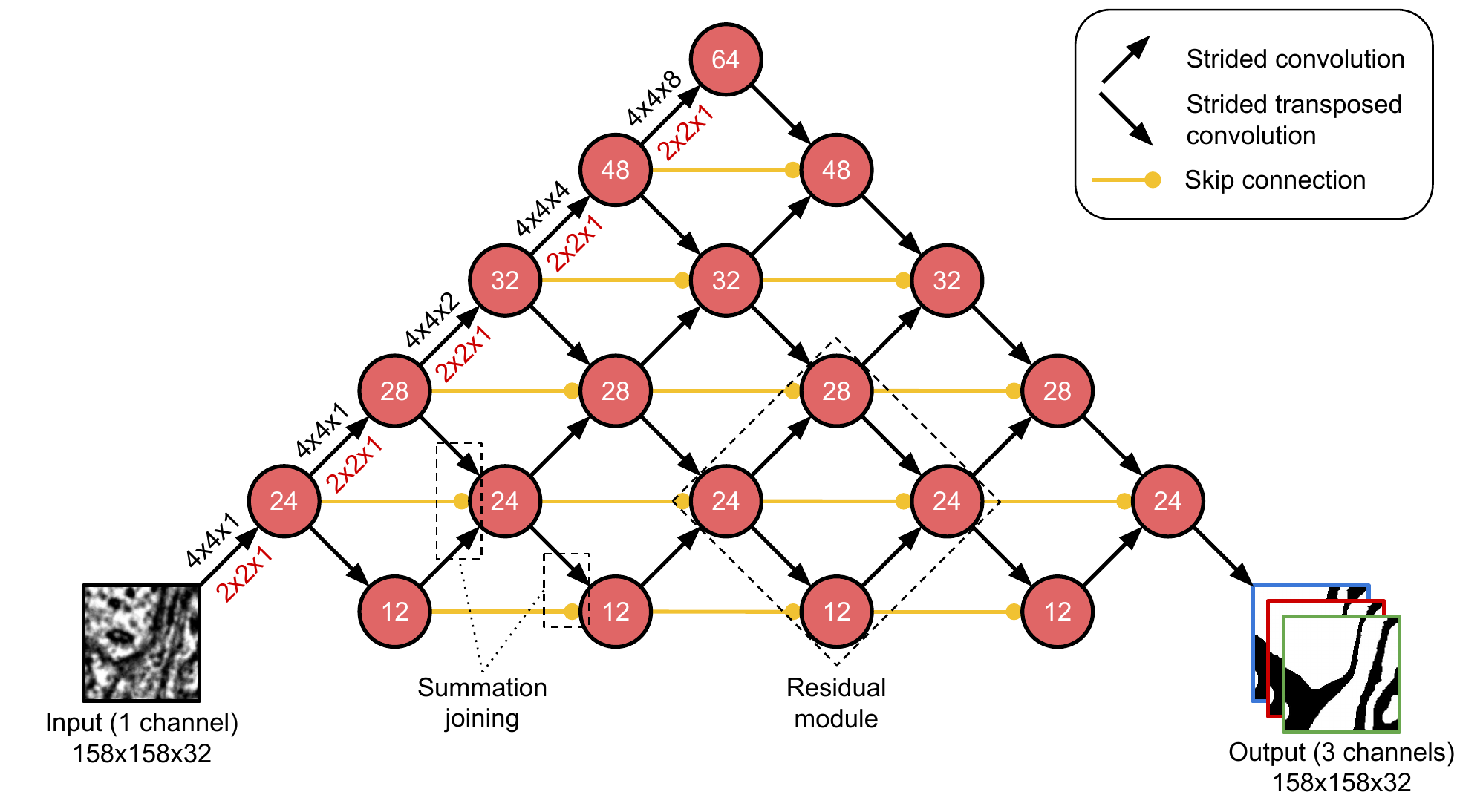}

\caption{Architecture for the baseline neuronal boundary detection. Each node
represents a layer and the number inside represents the number of feature maps.
The layers closer to the top of the diagram have lower resolution than the
layers near the bottom. The diagonal arrows represent strided convolutions,
while the horizontal arrows represent skip connections. Associated with the
diagonal arrows, the numbers indicate filter size (black) and strides (red) in
$x$, $y$, and $z$-dimension. The target for our boundary detection network is a
3D \emph{affinity graph}~\cite{boundary_detection,kisuk,funke2017deep}, thus
outputting three channels corresponding to $x$ (green), $y$ (red), and $z$
(blue) affinity maps, respectively.}

\label{fig:boundary_detector}
\end{figure}

Our proposed densely multiscale 3D convolutional network for neuronal boundary
detection is illustrated in Figure~\ref{fig:boundary_detector}. Our model is
built upon U-Net~\cite{unet} with several interesting architectural
augmentations. Our model can be viewed as a pyramidal stack of the basic
computational module (diamond-shaped box in Figure~\ref{fig:boundary_detector}).
This diamond-shaped module can be interpreted as a residual building block (see
Figure 2 in~\cite{resnet}) with two residual pathways, one top-down and the
other bottom-up. Thus our model is \emph{fully residual} in the sense that every
computational pathway involving horizontal information flow is passing through
the residual modules. Moreover, every residual module refines its input
representation by integrating both top-down and bottom-up information, thus
allowing for \emph{dense} intermixing of multiscale features. Our model's
\emph{dense} and \emph{fully residual} architecture allows an incremental and
iterative top-down/bottom-up refinement of internal representation, which is in
contrast to U-Net and variants' more restricted coarse-to-fine top-down
refinement~\cite{unet,pinheiro2016refine,lin2016pyramid}.

\begin{figure}[!t]
\centering
\includegraphics[width=0.65\linewidth]{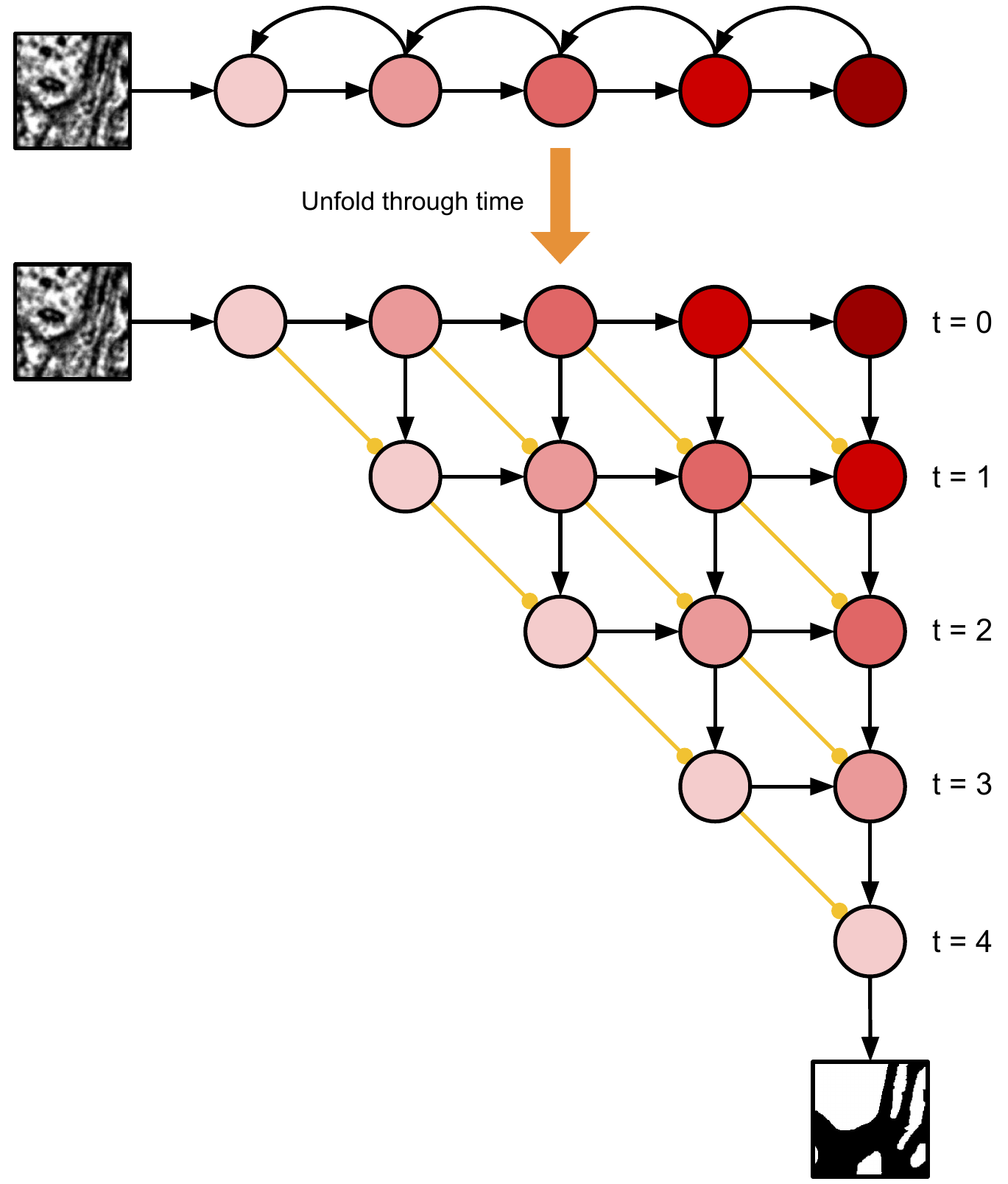}

\caption{Feedback recurrent convolutional network unrolled in time
(see Section~\ref{sec:deltanet} for details).}

\label{fig:unfold}
\end{figure}

From a different point of view, Figure~\ref{fig:unfold} illustrates another
important motivation for our densely multiscale convolutional net architecture.
Our model can be viewed as a feedback recurrent convolutional network unrolled
in time (Figure~\ref{fig:unfold}). Weight sharing across time makes our model
exactly equivalent to a convolutional net with recurrent feedback connections
unfolded through time, and this novel perspective provides a better framework
for understanding one of the unique characteristics of our model, i.e., the
incremental refinement of internal representation by interative integration of
top-down and bottom-up information. Our net's internal representation is
incremetally and iteratively refined over time by integrating the top-down
contextual information conveyed through the feedback recurrent connnections and
the higer spatial-frequency information relayed through the bottom-up
feedforward connections.

\begin{sidewaysfigure}[!t]
\centering
\includegraphics[width=1.0\linewidth]{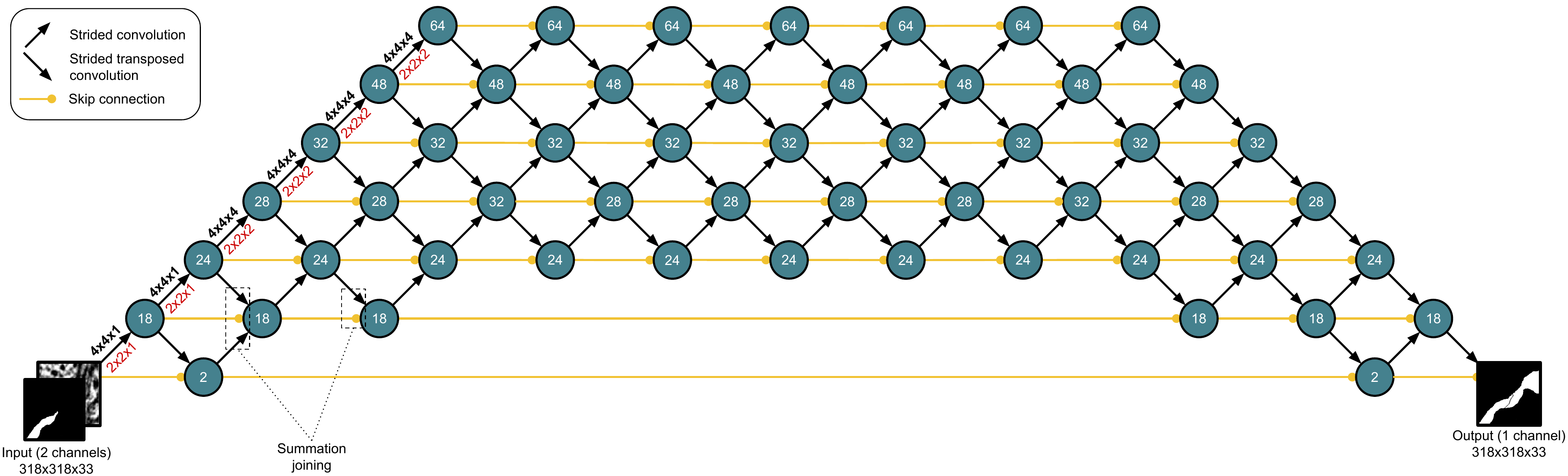}
\caption{Error-correcting network architecture.}
\label{fig:corrector}
\end{sidewaysfigure}

Note that we did not use weight sharing in the neuronal boundary detection net,
whereas we used weight sharing between some convolutions at the same height
in the error-detecting and error-correcting nets (Figure 2 and
Figure~\ref{fig:corrector}).

\paragraph{Architectural details} Due to the anisotropy of the resolution of the
images in our dataset, we design our networks so that the first convolutions
are exclusively 2D while later convolutions are 3D
(Figure~\ref{fig:boundary_detector}). The receptive field of a neuron in the
higher layers is therefore nearly isotropic and roughly cubic in physical size.
To limit the number of parameters in our model, we factorized all 3D
convolutions into a 2D convolution followed by a 1D convolution in
$z$-dimension. We employed exponential linear units (ELUs,~\cite{elu}) as
nonlinearity, except for the output layer with logistic activation functions. We
trained our nets to generate a 3D affinity
graph~\cite{boundary_detection,kisuk,funke2017deep}, thus outputting three
channels corresponding to $x$, $y$, and $z$-affinity map, respectively.

\subsection{Dataset}
Our dataset is a sample of mouse primary visual cortex (V1) acquired using
transmission electron microscopy at the Allen Institute for Brain Science. The
voxel resolution is $3.6~\text{nm} \times 3.6~\text{nm} \times 40~\text{nm}$.

Human experts produced multiple volumes of gold standard dense reconstruction,
in total $20$ volumes of size $512 \times 512 \times 100$. We trained our
boundary detector using $19$ volumes and used the last volume for training
validation. We applied the trained boundary detector on a new image volume of
size $2048 \times 2048 \times 100$ to obtain a preliminary segmentation, which
was then proofread by the experts to generate a bootstrapped ground truth
volume. This volume was used to optimize the parameters for watershed and mean
affinity agglomeration~\cite{kisuk}. Finally, the optimized segmentation
pipeline was applied to generate further bootstrapped ground truth for the error
detection and correction tasks.

\subsection{Training procedures} Our boundary detection networks were
implemented in the Caffe deep learning framework~\cite{jia2014caffe}. To train
our nets, we minimized the pixelwise binomial cross-entropy loss with
class-rebalancing using the Adam optimizer~\cite{adam}, initialized with
$\alpha=0.001$, $\beta_1=0.9$, $\beta_2=0.999$, and $\epsilon=0.01$. The network
weights were initialized following He et al.~\cite{he2015delving}. The learning
rate (or step size parameter $\alpha$ in the Adam optimizer) was halved when
validation loss plateaued out, five times in total at $35$K, $175$K, $250$K,
$300$K, and $480$K training iterations. We used a single patch of size
$158\times158\times32$ (i.e. minibatch of size 1) to compute gradients at each
training iteration. Each training sample was augmented with rotations,
reflections, warping, brightness and contrast perturbations, and simulated
misalignments~\cite{kisuk}. The training lasted for $800$K iterations until
convergence, which took about five days on a single NVIDIA Titan X Pascal GPU.

\subsection{Inference and postprocessing} We performed \emph{overlap-blending}
inference followed by watershed and mean affinity agglomeration~\cite{kisuk}. We
refer the interested readers to~\cite{kisuk} for further details.

%-------------------------------------------------------------------------------
% Section B. Network Architecture
%-------------------------------------------------------------------------------
\label{appendix:architecture}

\section{Per-object VI score}
\label{appendix:vi}
 Recall that the variation of information between two segmentations may be computed as
\begin{align*}
	VI_\text{split}&=-\frac 1 {\sum_{i,j} r_{ij}} \sum_{i,j} r_{ij} \log\left(r_{ij}/p_i\right),\\
	VI_\text{merge}&=-\frac 1 {\sum_{i,j} r_{ij}} \sum_{i,j} r_{ij} \log\left(r_{ij}/q_j\right),\\
	p_i&=\sum_j r_{ij},\\
	q_j&=\sum_i r_{ij},
\end{align*}
where $r_{ij}$ is the number of voxels in common between the $i^\text{th}$ segment of the ground truth segmentation and the $j^\text{th}$ segment of the proposed segmentation \cite{vi}.

We define the split and merge scores for ground truth segment $i$ as
\begin{align*}
	VI_\text{split}(i) &= -\sum_j r_{ij}/p_i \log(r_{ij}/p_i),\\
	VI_\text{merge}(i) &= -\sum_j r_{ij}/p_i \log(r_{ij}/q_j).
\end{align*}
Both quantities have units of nats. $VI_\text{split}(i)$ is zero if and only if
ground truth segment $i$ is contained within a segment in the proposed
segmentation, while $VI_\text{merge}(i)$ is zero if and only if ground truth
segment $i$ is the union of one or more segments in the proposed segmentation.
The total score $VI_\text{split}$ or $VI_\text{merge}$ is a weighted sum of the
per-object scores $VI_\text{split}(i)$, $VI_\text{merge}(i)$ respectively.

%-------------------------------------------------------------------------------
% Section B. Training Details
%-------------------------------------------------------------------------------
\section{Training details} The error-detecting and error-correcting networks
were implemented in TensorFlow \cite{tensorflow} and trained using 4 TitanX
Pascal GPUs with synchronous gradient descent. We used the Adam optimizer,
initialized with $\alpha=0.001$, $\beta_1=0.95$, $\beta_2=0.9995$, and
$\epsilon=0.1$ \cite{adam}.   Both nets were trained until the loss on a
validation set plateaued. The error-detecting net was trained for $700$K
iterations (approximately one week), while the error-correcting net was trained
for $1.7$M iterations (approximately three weeks).

\end{document}